\documentclass{article}

\usepackage[nonatbib]{neurips_2020}
\usepackage{amssymb}
\usepackage{amsmath}
\usepackage{bm,url,color,graphicx,cite,epstopdf,nicefrac,bbold, amsthm}
\usepackage{psfrag,subfigure,float}
\usepackage[algoruled,linesnumbered]{algorithm2e}
\usepackage{multirow}

\usepackage{algorithmic}
\usepackage[normalem]{ulem}
\usepackage{stmaryrd}
\usepackage{xcolor}
\usepackage{psfrag,framed,mdframed}
\usepackage{comment}
\usepackage{mdframed}
\usepackage{cleveref}
\usepackage{bibunits}
\usepackage[utf8]{inputenc}

\DeclareMathOperator*{\argmin}{arg\,min}
\usepackage{mathtools}

\definecolor{orange}{RGB}{255,107,0}
\definecolor{green}{RGB}{0,100,0}

\usepackage{steinmetz}

\renewcommand{\b}{\boldsymbol{b}}

\newcommand{\g}{\boldsymbol{g}}

\newcommand{\n}{\boldsymbol{n}}

\renewcommand{\u}{\boldsymbol{u}}

\newcommand{\y}{\boldsymbol{y}}
\newcommand{\x}{\boldsymbol{x}}
\newcommand{\z}{\boldsymbol{z}}

\newcommand{\bDelta}{\boldsymbol{\varDelta}}

\newcommand{\bdelta}{\boldsymbol{\delta}}

\newcommand{\wz}{\widehat{\boldsymbol{z}}}
\newcommand{\wu}{\widehat{\boldsymbol{u}}}
\newcommand{\wx}{\widehat{\boldsymbol{x}}}

\newcommand{\A}{\boldsymbol{A}}

\newcommand{\T}{{\!\top\!}}

\newcommand{\blambda}{\bm{\lambda}}

\newcommand{\cA}{\mathcal{A}}

\newcommand{\cC}{\mathcal{C}}

\newcommand{\cL}{\mathcal{L}}

\newcommand{\cN}{\mathcal{N}}

\newcommand{\cQ}{\mathcal{Q}}

\newcommand{\cV}{\mathcal{V}}

\newcommand{\bbR}{\mathbb{R}}

\newcommand{\bcC}{\bm \cC}

\DeclareMathOperator*{\minimize}{\textrm{minimize}}

\usepackage{xcolor}
\usepackage{psfrag,framed}
\usepackage{lipsum}
\usepackage{chapterbib}
\PassOptionsToPackage{normalem}{ulem}
\usepackage{ulem}
\definecolor{shadecolor}{RGB}{220,220,220}

\title{Communication-Efficient Distributed Asynchronous ADMM}

\author{%
  Sagar Shrestha \\
  Department of Electrical Engineering and Computer Science\\
  Oregon State University\\
  Corvallis, OR 97331 \\
  \texttt{shressag@oregonstate.edu} \\
}

\begin{document}

\maketitle

\begin{abstract}
In distributed optimization and federated learning, asynchronous \textit{alternating direction method of multipliers} (ADMM) serves as an attractive option for large-scale optimization, data privacy, straggler nodes and variety of objective functions. However, communication costs can become a major bottleneck when the nodes have limited communication budgets or when the data to be communicated is prohibitively large. In this work, we propose introducing coarse quantization to the data to be exchanged in aynchronous ADMM so as to reduce communication overhead for large-scale federated learning and distributed optimization applications. We experimentally verify the convergence of the proposed method for several distributed learning tasks, including neural networks.
\end{abstract}

\section{Introduction}
In federated learning, numerous connected devices seek to learn a common statistical model using but without sharing their private data. It naturally demands for distributed large-scale optimization methods. Therefore, ADMM is a attractive choice because of its ability to handle large classes of optimization problems including non-smooth objective \cite{boyd2011distributed}. However, the heterogeneous nature of the network encountered in federated learning makes it inefficient for synchronous updates as it does not fully utilize the node's computation and communication capabilities \cite{li2020federated}. As such, asynchronous ADMM has been proposed as a suitable method in \cite{chang2016asynchronous}, \cite{zhang2014asynchronous}.

While asynchronous ADMM appears to fix the issue of stragglers, a major hurdle in its application for large scale models is the communication bottleneck. When the data to be communicated is large, e.g., neural network parameters, or the communication resources are limited, e.g. battery operated devices, exacerbated by hundreds to millions of iterations needed for convergence, communication time can become a major bottleneck in the learning process. To this end, we propose to utilize coarse quantization of the data to be communicated for both exact and inexact asynchronous ADMM update. We show empirically that we are able to reduce around 90\% communication overhead for both exact and inexact asynchronous ADMM without any apparent loss in the convergence properties of the unquantized version.

\section{Related Works}
Asynchronous ADMM for inexact primal updates was proposed in \cite{hong2014distributed} and for exact updates in \cite{zhang2014asynchronous, chang2016asynchronous}. However, communication cost reduction was not considered in those references. Recently, \cite{ryu2021differentially, yue2021inexact, zhou2021communication} considered inexact ADMM for the synchronous case in the context of federated learning. Although \cite{zhou2021communication} claim communication efficiency of their method, the gain is only derived from reduction in communication round from multiple local updates. However, communication requirement in each outer iteration can still be prohibitive for large scale problems. In addition to this effort, the proposed method in this project significantly reduces the communication overhead of each round. Further, the aforementioned references only provide simulations for convex problems (e.g., linear regression, logistic regression, LASSO, etc). In contrast, we provide simulations validating the proposed method in the case of deep neural networks, which is more representative of federated learning applications. 

Closely related to ADMM, \cite{chen2021communication} proposed a primal-dual method utilizing compression and error-feedback to achieve communication efficiency in distributed optimization. However, their method is limited to the synchronous case and merely deals with the uplink communication overhead. In contrast, the proposed method in this projects reduces both the uplink and downlink communication overhead and incorporates the asynchronous update case, which is essential to dealing with stragglers in hetergeneous networks. 

\section{Background}
In this section, we present the distributed asynchronous ADMM algorithm to provide necessary background for the proposed communication-efficient method.  
\subsection{ADMM}
We are interested in solving the following optimization problem:
\begin{equation}\label{eq:objective}
\minimize_{\x \in \bbR^M}  f(\x) + h(\x),
\end{equation}
where $f:\bbR^M \to \bbR$ is a smooth cost function, and $h: \bbR^M \to \bbR$ is a convex (possibly non-smooth) regularization term. To solve the problem using N agents in a distributed setting, we require that $f(\x)$ be decomposed into $N$ local objectives:
$$ f(\x) = \sum_{i=1}^N f_i(\x),$$
where $f_i:\bbR^M \to \bbR$ is a local objective function which only depends upon the data at node $i$. This allows us to reformulate problem \eqref{eq:objective} as a global variable consensus optimization problem as follows:
\begin{align}\label{eq:distributed_obj}
\minimize_{\x_1, \dots, \x_N, \z} &~ \sum_{i=1}^N f_i(\x_i)  + h(\z) \nonumber \\
\text{subject to} &~ \x_i = \z, ~ i=1,2, \dots, N, 
\end{align}
where $\z$ is the consensus variable, and $\x_i$ is the node $i$'s proxy for the global variable $\z$. Essentially, we wish to distribute the optimization of $f(\cdot)$ to $N$ nodes, where each node optimizes for its local objective $f_i(\cdot)$. This results in an iterative optimization procedure such that in each iteration the nodes solve their corresponding local sub-problems, and communicate the solution to a central controller or the server. The server collects the local variables, updates the consensus variable and distributes it to all the nodes. 

\begin{figure}[t]
    \centering
    \includegraphics[width=0.2\linewidth]{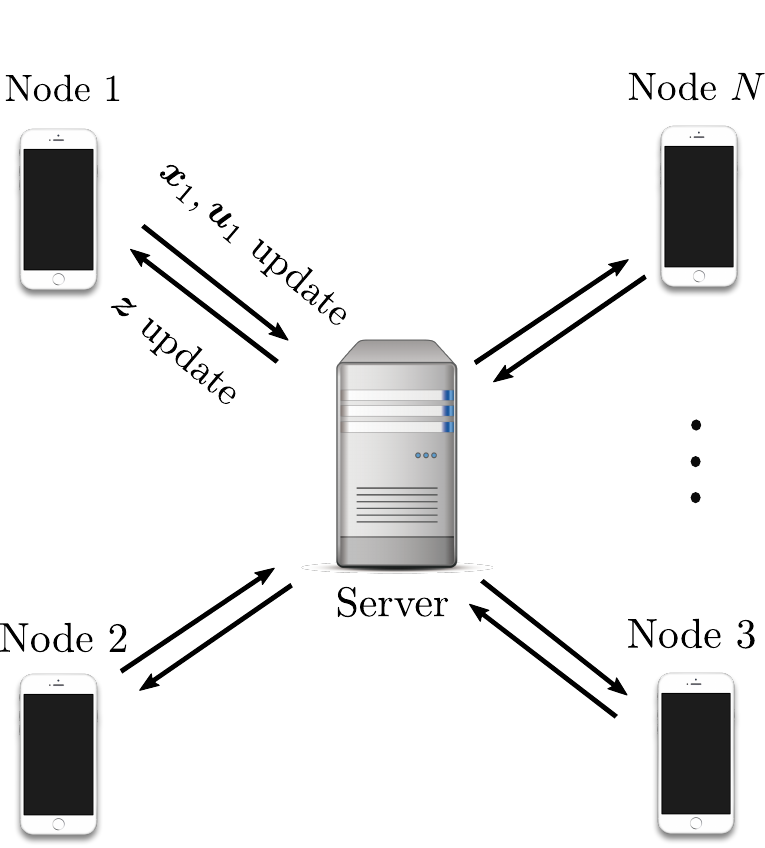}
    \caption{Illustration of Distributed Optimization using ADMM.}
    \label{fig:distr_setting}
\end{figure}

ADMM combines the features of dual ascent and augmented Lagrangian method to solve problem \eqref{eq:distributed_obj}. The augmented Lagrangian of \eqref{eq:distributed_obj} can be written as follows:
\begin{equation}
    \cL\left(\{\x_i\}_{i=1}^N, \z, \{\blambda_i\}_{i=1}^N\right) = \sum_{i=1}^N f_i(\x_i)  + h(\z) + \blambda_i^\T(\x_i - \z) + \frac{\rho}{2} \sum_{i=1}^N \| \x_i - \z\|_2^2,
\end{equation}
where $\blambda_i \in \bbR^M$ is the dual variable associated with the $i$th constraint and $\rho> 0$ is the penalty parameter. The above can be simplified as:
\begin{equation}\label{eq:alm}
    \cL\left(\x, \z, \u\right) = \sum_{i=1}^N f_i(\x_i)  + h(\z) + \frac{\rho}{2} \sum_{i=1}^N \| \x_i - \z + \u_i\|_2^2,
\end{equation}
where $\x = [\x_1, \dots, \x_N]^\T$, $\u = [\u_1, \dots, \u_N]^\T$, and $\u_i = \nicefrac{\blambda_i}{\rho}$. ADMM involves iteratively optimizing all the primal variables,  $\{\x_i\}_{i=1}^N$ and $\z$, followed by one step gradient ascent of the dual variables in a Gauss-Seidel fashion \cite{boyd2011distributed}. Figure \ref{fig:distr_setting} illustrates the distributed ADMM scenario. The nodes and the server are connected in a star topology and communicate local and consensus variables with each other. Following steps summarize one step of ADMM update:
\begin{subequations}\label{eq:admm_update}
\begin{align}
    \x_i^{(r+1)} &\leftarrow \argmin_{\x_i \in \bbR^M} ~ f_i(\x_i) + \frac{\rho}{2} \| \x_i - \z^{(r)} + \u_i^{(r)}\|_2^2, ~ i = 1, \dots, N \label{eq:admm_x_update}\\
    \z^{(r+1)} &\leftarrow \argmin_{\z \in \bbR^M} ~ h(\z) + \frac{\rho}{2} \sum_{i=1}^N \| \x_i^{(r+1)} - \z + \u_i^{(r)}\|_2^2 \label{eq:admm_z_update} \\
    \u_i^{(r+1)} &\leftarrow \u_i^{(r)} + (\x_i^{(r+1)} - \z^{(r+1)}), ~ i = 1, \dots, N, \label{eq:admm_u_update} 
\end{align}
\end{subequations}
where $r$ is the current iteration index.
Note that the updates \eqref{eq:admm_x_update} and \eqref{eq:admm_u_update} can be performed locally at the nodes and only requires the consensus variable. \eqref{eq:admm_z_update}, on the other hand, requires all local variables: $\{\x_i \}_{i=1}^N, \{ \u_i\}_{i=1}^N$, and is carried out at the server. Therefore, the updates at node $i$ are as follows:
\begin{subequations}\label{eq:node_update}
\begin{align}
   & \x_i^{(r+1)} \leftarrow \argmin_{\x_i \in \bbR^M} ~ f_i(\x_i) + \frac{\rho}{2} \| \x_i - \z^{(r)} + \u_i^{(r)}\|_2^2, ~ i = 1, \dots, N \label{eq:node_x_update}\\
   & \u_i^{(r+1)} \leftarrow \u_i^{(r)} + (\x_i^{(r+1)} - \z^{(r)}), ~ i = 1, \dots, N. \label{eq:node_u_update} \\
    & \text{Send } \x_i^{(r+1)}, \u_i^{(r+1)} \text{ to the server}. \nonumber 
\end{align}
\end{subequations}

Similarly the server update is as follows:
\begin{subequations}\label{eq:server_update}
\begin{align}
    & \z^{(r+1)} \leftarrow \argmin_{\z \in \bbR^M} ~ h(\z) + \frac{\rho}{2} \sum_{i=1}^N \| \x_i^{(r+1)} - \z + \u_i^{(r+1)}\|_2^2 \label{eq:server_z_update}. \\
    & \text{Broadcast } \z^{(r+1)} \text{ to the nodes}. \nonumber 
\end{align}
\end{subequations} 

\subsection{Asynchronous ADMM}
The distributed implementation in \eqref{eq:node_update} and \eqref{eq:server_update} can be viewed as synchronous ADMM because all the nodes are synchronized, i.e., all nodes complete one update before the server makes its update. Therefore this distributed implementation retains the same convergence properties of the undistributed version. However, the speed of such method is limited to the speed of the slowest node. This does not fully utilize the computation abilities of the faster nodes (one with more computation and/or communication resources) in the network. 

To address the limitation of synchronous setting, asynchronous ADMM has been considered for exact updates of primal variables in \cite{chang2016asynchronous, zhang2014asynchronous} and inexact updates in \cite{hong2014distributed}. In the asynchronous setting, the server does not wait for all the nodes to complete their updates. Instead the server performs its computation using updates from a subset of nodes. Specifically, let $P$ be the minimum number of nodes that can trigger a server update. Then during iteration $r$, let $\cA_r \subseteq \cV = \{ 1, \dots, N\}$ be the set of nodes that have completed its operation. The server waits until $|\cA_r| \geq P$ before performing the server update. In addition, in order to ensure that updates from all nodes arrive within a limited time frame, we define $\tau$ as the maximum delay (in iterations) allowed for the arrival of any node. This means that in any iteration, the server waits for the nodes that have not updated for $\tau-1$ iterations.

With this, we can write the the node update as follows:
\begin{subequations}
\begin{align}
   & \x_i^{(r+1)} \leftarrow \begin{cases} \argmin_{\x_i } ~ f_i(\x_i) + \frac{\rho}{2} \| \x_i - \z^{(r)} + \u_i^{(r)}\|_2^2,  \quad \forall i \in \cA_r \\
   \x_i^{(r)} \quad \forall i  \in \cV \backslash \cA_r \end{cases}\\
   & \u_i^{(r+1)} \leftarrow \begin{cases} \u_i^{(r)} + (\x_i^{(r+1)} - \z^{(r)}), \quad \forall i \in \cA_r \\
   \u_i^{(r)}, \quad \forall i  \in \cV \backslash \cA_r
   \end{cases} \\
    & \text{Send } \x_i^{(r+1)}, \u_i^{(r+1)}. \nonumber 
\end{align}
\end{subequations}
The server update is the same as \eqref{eq:server_update}. 

\begin{figure}
    \centering
    \includegraphics[width=\linewidth]{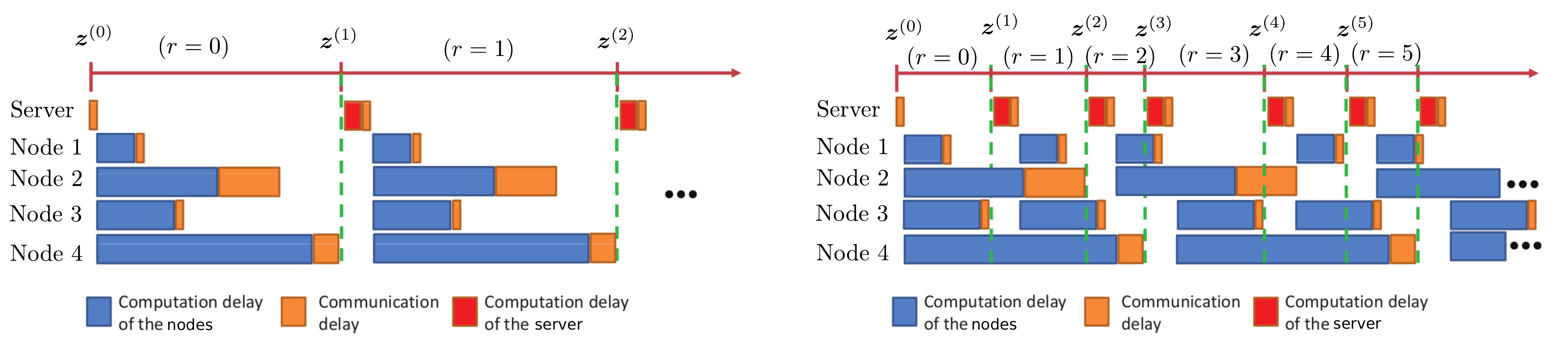}
    \caption{Illustration of synchronous and asynchronous distributed ADMM updates for $P=2$. (Figure reproduced from \cite{chang2016asynchronous}). }
    \label{fig:async_illus}
\end{figure}

Figure \ref{fig:async_illus} illustrates the difference between the synchronous and asynchronous ADMM for $P=2$. We can see that in the synchronous update the server waits for all the nodes to finish computation and communication before starting its update. Whereas, in the asynchronous update, the server starts its computation after receiving updates from $P$ nodes. 

\section{Communication-Efficient Asynchronous ADMM}
Asynchronous ADMM allows utilization of computation and/or communication resources at the nodes. However, when $M$ is large (e.g., in neural networks, M can be in the order of millions), the communication cost at each round can be prohibitive, not to mention the number of iterations required for convergence of the algorithm. For instance, if $M=10,000,000$, each node needs to upload $~640$MB of data at each iteration. This is not feasible if the node is a mobile device or the communication frequency is large. Therefore, it is well-motivated to consider further communication reduction per each update. To this end, we propose a carefully designed communication reduction technique for asynchronous distributed ADMM. 

\subsection{Compression and Error-Feedback}
We utilize the ideas from gradient compression literature \cite{ basu2019qsparse, bernstein2018signsgd, karimireddy2019error} to reduce the communication cost of the exchanged information. Specifically, we introduce a compression operator, $\bcC:\bbR^M \to \cQ^M$. Here $\cQ$ is the quantized domain, a subset of real valued domain. The idea is that 
$$\bcC(\y) \approx \y,$$
but $\bcC(\y)$ requires much fewer number of bits to represent than $\y$. The compressor can be quantization based \cite{bernstein2018signsgd, alistarh2017qsgd} or sparsification based \cite{basu2019qsparse, stich2018sparsified}. Moreover, we note that instead of communicating $\bcC(\y^{(r)})$ in iteration $r$, we can reduce the compression error by communicating $\bcC(\y^{(r+1)} - \y^{(r)})$. This is because the change in the iterate, $\y^{(r+1)} - \y^{(r)}$, is supposed to converge to zero for any converging algorithm.  

In addition to compressing the information to be exchanged, we utilize error-feedback method introduced in \cite{karimireddy2019error} that has been shown to be important to ensure convergence in some compressor (e.g.,  \cite{bernstein2018signsgd}) or improve convergence rate in others (e.g., \cite{alistarh2017qsgd}). To explain error-feedback,  let $\y^{(r)}$ be the iterate we want to communicate from a source to a destination. Let the iterate be obtained at the source by the following relation:
$$ \y^{(r+1)} \leftarrow \y^{(r)} + \g^{(r)},$$
where $\g^{(r)}$ is the change in the iterate in iteration $r$. Let $\widehat{\y}^{(r)}$ be the estimate of $\y^{(r)}$ at the destination, which is obtained as follows:
\begin{align*}
\widehat{\y}^{(r+1)} &\leftarrow \widehat{\y}^{(r)} + \bcC(\g^{(r)}) \\
                   & = \widehat{\y}^{(r)} + \g^{(r)} + \bdelta^{(r)} \\
                   & = \y^{(0)} + \sum_{t=0}^{r} \g^{(t)} + \sum_{t=0}^{r} \bdelta^{(t)} \\
                   & = \y^{(r+1)}  + \underbrace{\sum_{t=0}^{r} \bdelta^{(t)}}_{\text{aggregated error}},
\end{align*}
where $\bdelta^{(r)}$ is the compression error in iteration $r$. We can see that the error in each iteration keeps integrating. Therefore when $r$ becomes large, $\widehat{\y}^{(r+1)}$ can be very far from $\y^{(r)}$. In error-feedback, we \textit{feedback} the error of previous iteration, $\bdelta^{(r-1)}$, along with the $\g^{(r)}$. Therefore, the source transmits $\bcC(\g^{(r)} -\bdelta^{(r-1)})$. This results in the following update of the estimate at the destination:
\begin{align*}
\widehat{\y}^{(r+1)} &\leftarrow \widehat{\y}^{(r)} + \bcC(\g^{(r)} - \bdelta^{(r-1)}) \\
                   & = \widehat{\y}^{(r)} + \g^{(r)} - \bdelta^{(r-1)} + \bdelta^{(r)} \\
                   & = \y^{(0)} + \sum_{t=0}^{r} \g^{(t)} + \sum_{t=0}^{r}(\bdelta^{(t)} - \bdelta^{(t-1)}) \\
                   & = \y^{(r+1)}  + \bdelta^{(r)}.
\end{align*}
This shows that we have eliminated the error terms from previous iterations ensuring that $\widehat{\y}^{(r)}$ is close to $\y^{(r)}$ given sufficiently accurate compressor. 

\subsection{Asynchronous ADMM with Compression and Error-Feedback}
Now, let us apply compression along with error feedback to asynchronous ADMM. Let $\widehat{\x}_i^{(r)}$ and  $\widehat{\u}_i^{(r)}$ be the server's estimates of $\x_i^{(r)}$ and $\u_i^{(r)}$ in iteration $r$. Similarly, let $\widehat{\z}^{(r)}$ be the nodes' estimate of $\z^{(r)}$. Then the node and server operations are carried out as follows:

\textbf{Node Operations.} In iteration $r$, node $i$ updates its local variables using its estimate of the $\z^{(r)}$, given by $\widehat{\z}^{(r)}$, as follows:
\begin{subequations}
\begin{align}
   & \x_i^{(r+1)} \leftarrow \begin{cases} \argmin_{\x_i } ~ f_i(\x_i) + \frac{\rho}{2} \| \x_i - \wz^{(r)} + \u_i^{(r)}\|_2^2,  \quad \text{if } i \in \cA_r \label{eq:qadmm_x_update} \\
   \x_i^{(r)} \quad \text{if }  i  \in \cV \backslash \cA_r \end{cases}\\
   & \u_i^{(r+1)} \leftarrow \begin{cases} \u_i^{(r)} + (\x_i^{(r+1)} - \wz^{(r)}), \quad \text{if }  i \in \cA_r \\
   \u_i^{(r)}, \quad \text{if }  i  \in \cV \backslash \cA_r
   \end{cases} 
\end{align}
\end{subequations}
Node $i$ computes the data to be communicated to the server, $\bDelta_{\x_i}^{(r)}$ and $\bDelta_{\u_i}^{(r)}$, change of the iterates along with the error from previous iteration as follows:
\begin{align}
    \bDelta_{\x_i}^{(r)}  &=  \underbrace{\x_i^{(r+1)} - \x_i^{(r)}}_{\text{current change}} + \underbrace{\x_i^{(r)} - \wx_i^{(r)}}_{\text{previous error}} = \x_i^{(r+1)} - \wx_i^{(r)} \\
    \bDelta_{\u_i}^{(r)}  &=   \u_i^{(r+1)} - \wu_i^{(r)} \\
\end{align}
Node $i$ then sends $\bcC(\bDelta_{\x_i}^{(r)})$ and $\bcC(\bDelta_{\u_i}^{(r)})$ to the server. The server then updates its estimate of node $i$'s local variables as follows:
\begin{align}
    \wx_i^{(r+1)} \leftarrow \wx_i^{(r)} + \bcC(\bDelta_{\x_i}^{(r)}), \\
    \wu_i^{(r+1)} \leftarrow \wu_i^{(r)} + \bcC(\bDelta_{\u_i}^{(r)}).
\end{align}
Note that above operation is also carried out at the nodes as they requires $\wx_i^{(r+1)}$ to compute the error term (for error feedback) in the next iteration.

\textbf{Server Operations.} 
The server updates $\z$ with the estimates of the node variables as follows:

\begin{align}
    & \z^{(r+1)} \leftarrow \argmin_{\z \in \bbR^M} ~ h(\z) + \frac{\rho}{2} \sum_{i=1}^N \| \wx_i^{(r+1)} - \z + \wu_i^{(r+1)}\|_2^2. \label{eq:server_z_update}. 
\end{align}
Similar to the operation in the nodes, the server computes $\bDelta_{\z}^{(r)}$ as $\bDelta_{\z}^{(r)} = \z^{(r+1)} - \wz^{(r)}.$
The server then broadcasts $\bcC(\bDelta_{\z}^{(r)})$ to all the nodes. The nodes and the server update the node's estimate of $\z$ as follows:
\begin{align}
    \wz^{(r+1)} \leftarrow \wz^{(r)} + \bcC(\bDelta_{\z}^{(r)}).
\end{align}

The algorithm is referred to as \textit{Quantized} ADMM (QADMM) and summarized in Algorithm \ref{algo:qadmm}. Note that we use a subroutine, \texttt{simulate-async}(), in order to simulate the asynchronous scenario. Specifically, we assume that \texttt{simulate-async}() is an oracle that provides us the set of nodes that will complete their computation and communication within the next iteration.

\textbf{Choice of Compressor. } In the simulations, we use a random compressor introduced in \cite{alistarh2017qsgd}, which allows for multi-precision quantization and has favourable convergence properties in practice. 

To quantize $\bDelta \in \bbR^{M}$, $\bDelta \not = \mathbf{0}$, ${\bm \cC}(\bDelta)$ is computed as follows. 
We first divide the range from $0$ to $1$ into $S$ intervals of equal width, where $S$ is related to the number of levels of quantization. Specifically, if $q$ is the number of bits used to represent a scalar value, $S = 2^{q-1} -1$. The quantization operation operates elementwise. Therefore, for each element $\bDelta(m)$, we can find an interval $[p/S, (p+1)/S], p\in\{0,\ldots,S-1\}$, such that the normalized value $|\bDelta(j, k)| / \|\bDelta\|_{\rm max} \in [p/S, (p+1)/S]$. 
Next, a Bernoulli random variable, $h(\bDelta(j, k), S)$, is defined as follows:
\begin{equation}\label{eq:compression_prob}
	h(\bDelta(j, k), S) = \begin{cases}
		p/S \quad \text{w.p.} \quad 1 - \left( \frac{|\bDelta(j, k)|}{ \|\bDelta\|_{\rm max}}S - p \right) \\
		(p+1)/S \quad \text{otherwise.}
	\end{cases} 
\end{equation}
Finally, we unnormalize $h(\bDelta(j, k), S)$ using the sign and magnitude information as $[{\bm \cC}(\bDelta)]_{m}	= \|\bDelta\|_{\rm max} {\rm sgn}(\bDelta(m)) \cdot h(\bDelta(m), S),$
where ${\rm sgn}(\cdot)$ is the sign operator.

\section{Simulation Results}
In this section, We present simulations to validate the effectiveness of the proposed method. 

\subsection{LASSO}
Consider the LASSO problem 
\begin{align}\label{eq:lasso}
    \minimize_{\x \in \bbR^M} \sum_{i=1}^N \|\A_i \x - \b_i \|_2^2 + \theta \|\x \|_1,
\end{align}
where $\A_i \in \bbR^{H \times M}$, $\b_i \in \bbR^{H}$, $\theta > 0$.

We reformulate \eqref{eq:lasso} to fit ADMM based distributed optimizatiion framework as follows:
\begin{align*}
    \minimize_{\x_1, \dots, \x_N, \z} ~ & \sum_{i=1}^N \|\A_i \x_i - \b_i \|_2^2 + \theta \|\z \|_1 \\
    \text{subject to } ~& \x_i = \z \quad i=1,\dots, N.
\end{align*}
We can see that $f_i(\x_i) = \|\A_i \x_i - \b_i \|_2^2.$
Therefore, each node operation only depends upon its local data $\A_i$ and $\b_i$. The primal update in \eqref{eq:qadmm_x_update} is a least square problem which is solved optimally at each iteration. Similarly the consensus update in \eqref{eq:server_z_update} is a proximal update with respect to L-1 norm. The optimal solution is given by soft-thresholding operation. Therefore, this is an example of exact minimization based QADMM. 

We generate the elements of $\A_i$ using standard normal distribution, i.e., $\sim \cN(0,1)$. $\b_i$'s are generated following $\b_i = \A_i \z_0 + \n_i$, where $\z_0  \in \bbR^M$ is a sparse random vector with $0.2M$ non-zero elements sampled from $\cN(0,1)$, and $\n_i$ is noise vector sampled from $\cN(0,0.01)$. We use $N=16$. To simulate the asynchronous case, we implement \texttt{simulate-async}$()$ subroutine in Algorithm \ref{algo:qadmm} by the following procedure. We randomly split $N$ nodes into two sets. For each element of the first set, we set the probability of getting selected as 0.1 and for the other half, the probability is set to 0.8. Thus the nodes selected by \texttt{simulate-async}() complete their local updates and communicate them to the server within the next iteration.

\textbf{Metric.} To measure the progress of the algorithm towards the optimal, we observe the accuracy defined as follows:
\begin{align}
    {\rm Accuracy}(r) = \frac{|\cL(\x^{(r)}, \z^{(r)}, \u^{(r)}) - F^\star|}{F^\star}
\end{align}
where $\cL(\cdot)$ is the augmented lagrangian in \eqref{eq:alm}, and $F^\star$ is the optimal objective value for \eqref{eq:lasso} In order to evaluate the gain in communication efficiency, we define communication bits as follows:
\begin{align}
    {\rm Communication ~bits} = \frac{{\rm total~ bits~ communicated~ between~ the~ nodes~ and~ the ~server }}{M}
\end{align}

\textbf{Baseline.} We compare \texttt{QADMM} with its unquantized version referred to as \texttt{async ADMM}.

\begin{figure}[t]
    \centering
    \includegraphics[width=0.8\linewidth]{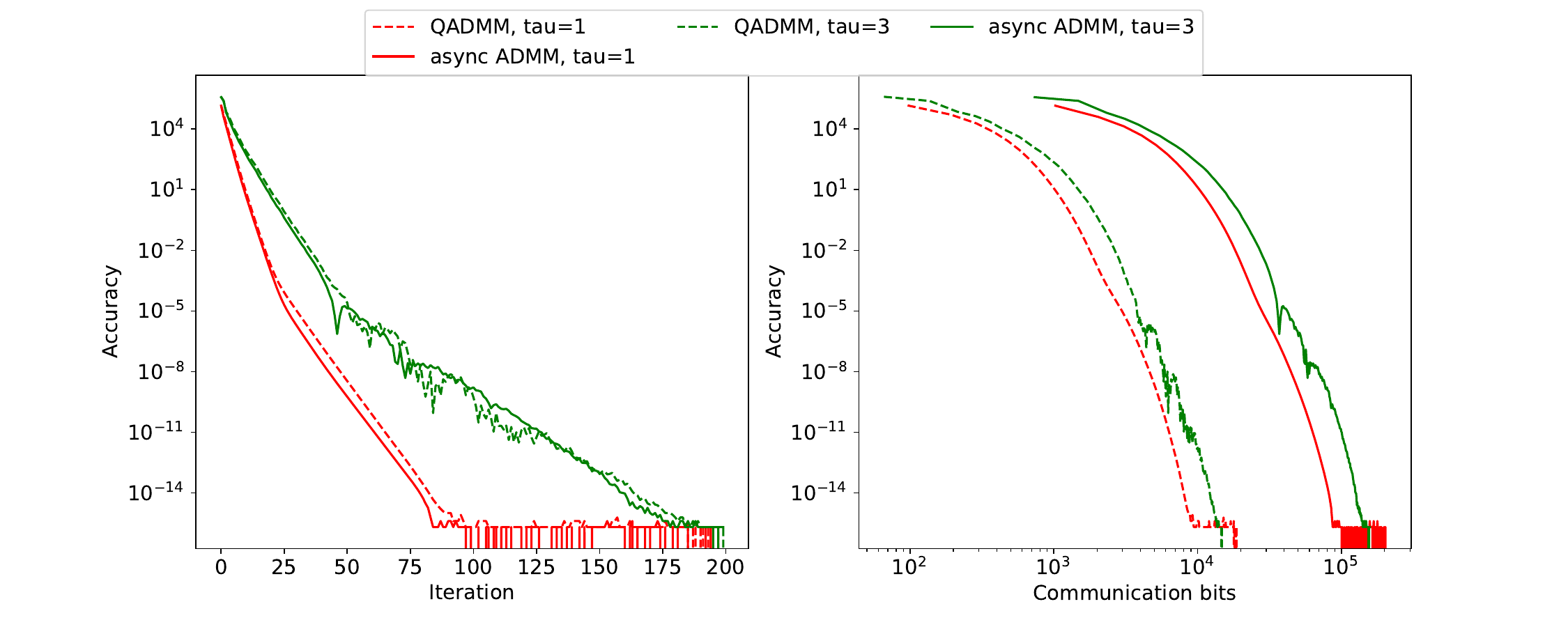}
    \caption{Classification accuracy attained by the proposed method, \texttt{QADMM}, with respect to the unquantized version for various $\tau$ vs. iterations and communication bits.}
    \label{fig:lasso}
\end{figure}

Figure \ref{fig:lasso} shows the accuracy with respect to iteration and communication bits for the case when $(M, \rho, \theta, N, H)=(200, 500, 0.1, 16, 100) $ for $\tau \in \{1,3\}$. The result is averaged over 10 Monte Carlo trials. Note that $\tau=1$ corresponds to the synchronous case. We use $q=3$ bits per scalar for both the downlink and uplink compression. We can see that there is apparently no degradation in convergence behavior due to compression. Further, the plot with respect to the communication bits shows that \texttt{QADMM} approaches the optimal much faster ($\approx 10 \times$) than its unquantized version. Specifically, we observe that to obtain an accuracy of $10^{-10}$, \texttt{QADMM} requires $90.62\%$ less communication bits than the unquantized version.   

\begin{figure}[t]
    \centering
    \includegraphics[width=.8\linewidth]{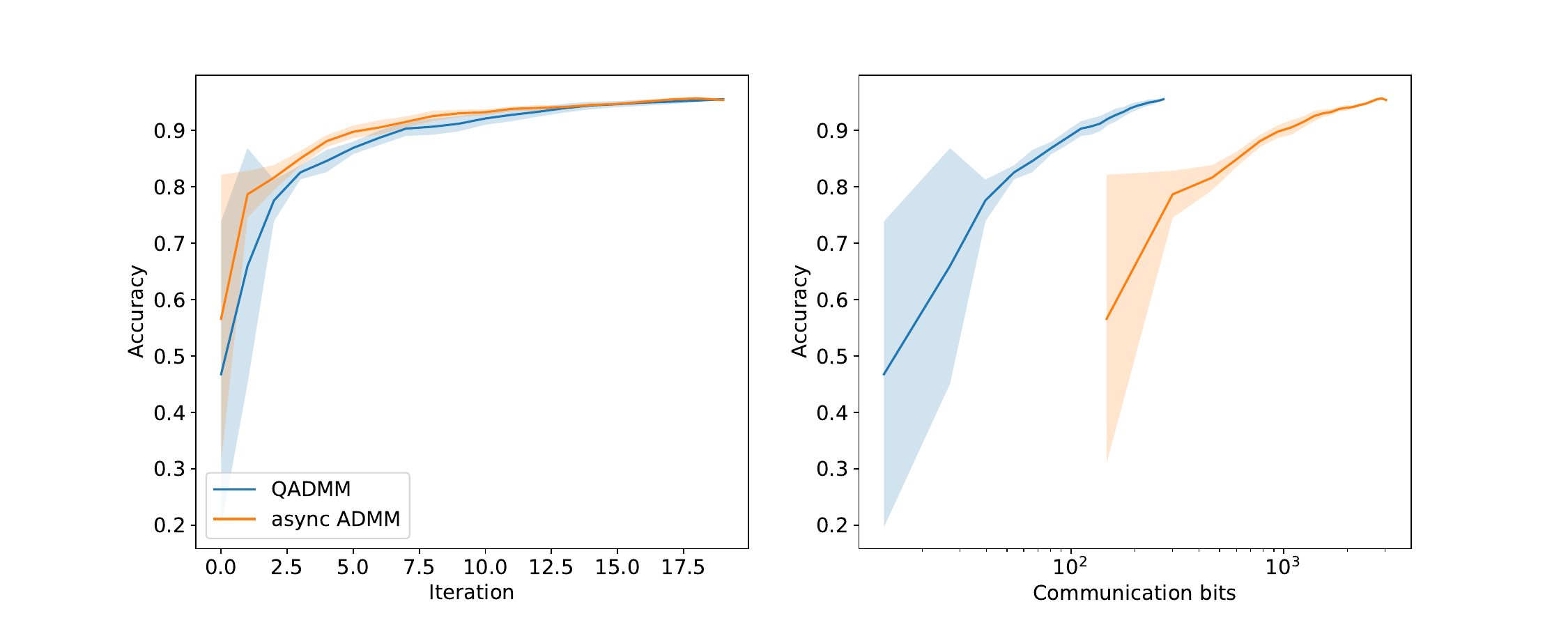}
    \caption{Classification accuracy attained by the proposed method, \texttt{QADMM}, with respect to the unquantized version vs. iterations and communication bits. }
    \label{fig:mnist}
\end{figure}

\subsection{MNIST Classifier}
In this subsection, we validate the effectiveness of the proposed method for neural-network based classifiers. Since many of the recent applications of federated learning employ neural network based models, this example serves to validate the usefulness of \texttt{QADMM} in real-world applications.

We consider a \textit{convolutional neural network} (CNN) based classifier with 6 layers, where the first 5 layers are convolutional layers and the final layer is fully connected. The filter size for the first 5 layers is $3 \times 3$, stride of 2, padding of 1, and number of filters  are 16, 32, 64, 128, and 128 respectively. The final fully connected layer has 10 neurons to classify among the 10 digits with a sigmoid activation at the output. The total number of parameters $M = 246,762$. Note that we cannot exactly solve the primal update in \eqref{eq:qadmm_x_update} in this case because of the highly non-convex nature of the problem. Therefore, we return inexact solution by running 10 iterations of gradient descent with a batch size of 64 at each update. We use ADAM \cite{kingma2014adam} with an initial learning rate of $0.001$ for the inexact primal update. Consequently, this is an example of inexact asynchronous ADMM.

We use $N=3$, and randomly divide the $60,000$ training examples into $N$ partitions.  Similar to the previous example, for each call to the subroutine \texttt{simulate-async}(), we create two groups of nodes with each node independently assigned to either of the groups with equal probability. The nodes in the first group have 0.1 probability of being selected by the subroutine and those in the second group have 0.8 probability of being selected. We set the number of bits $q=3$ and $\tau=3$.

Figure \ref{fig:mnist} shows the classification accuracy attained by \texttt{QADMM} and \texttt{async ADMM} with respect to iterations and communication cost on a held out test set of size 10,000. The result is averaged over 5 Monte Carlo trials. We can see that there is virtually no degradation in convergence properties of the proposed method relative to the unquantized version. However, there is a significant reduction in communication bits required to reach the desired classification accuracy. Specifically, to attain a classification accuracy of 95\%, \texttt{QADMM} requires 91.02\% less communication bits than the unquantized version.

\section{Conclusion}
In this project, we proposed a communication-efficient method for asynchronous distributed implementation of exact and inexact ADMM. With carefully designed compression scheme and error feedback, we showed experimentally that the proposed method can reduce communication cost by more than 90\% in both uplink and downlink direction for both convex problems as well as non-convex problem involving deep neural networks.  

\clearpage
 \bibliographystyle{IEEEtran}
 \bibliography{main}
 
\clearpage

\begin{algorithm}[t]
\label{algo:qadmm}
\footnotesize
\SetAlgoLined

\tcp{Initialization at the nodes}
\For{$i \leftarrow 1:N$}{
    Initialize $\x_i^{(0)}, \u_i^{(0)}$\;
    Transmit $\x_i^{(0)}, \u_i^{(0)}$ to the server using full precision (e.g., 32-bits per scalar)\;
}

\tcp{Initialization at the server}
$d_1 = \cdots = d_N = 0$\;
$\wx_i^{(0)} \leftarrow \x_i^{(0)}, \quad \wu_i^{(0)} \leftarrow \u_i^{(0)}, \forall i$\; \tcp*{received with full precision}
$ \z^{(0)} \leftarrow \argmin_{\z} ~ h(\z) + \frac{\rho}{2} \sum_{i=1}^N \| \wx_i^{(0)} - \z + \wu_i^{(0)}\|_2^2$\;
Broadcast $\z^{(0)}$ to the nodes using full precision\;
$r \leftarrow 0$\;

\While{some stopping criteria is not met}{
    \tcp{At the nodes}
    \For{ $i=1:N$ in parallel}{
        \eIf{$r = 0$}{
            $\wz^{(0)} \leftarrow \z^{(0)}$\; \tcp*{received with full precision}
        }{
            Receiving $\bcC(\bDelta_{\z}^{(r-1)})$ from the server\;
            $\wz^{(r)} \leftarrow \wz^{(r-1)} + \bcC(\bDelta_{\z}^{(r-1)})$\;
        }
        \eIf{node $i \in \cA_r$}{
            $\x_i^{(r+1)} \leftarrow \argmin_{\x_i } ~ f_i(\x_i) + \frac{\rho}{2} \| \x_i - \wz^{(r)} + \u_i^{(r)}\|_2^2$\;
            $\u_i^{(r+1)} \leftarrow \u_i^{(r)} + (\x_i^{(r+1)} - \wz^{(r)})$\;
            Send $\{\bcC( \bDelta_{\x_i}^{(r)}), \bcC( \bDelta_{\u_i}^{(r)})\}$ to the server\;
        }{
            $\x_i^{(r+1)} \leftarrow \x_i^{(r)}$\;
            $\u_i^{(r+1)} \leftarrow \u_i^{(r)}$\;
        }
    }
    \tcp{At the Server}
    Receive $\{\bcC( \bDelta_{\x_i}^{(r)}), \bcC( \bDelta_{\u_i}^{(r)})\}_{ i \in \cA_r}$ from the nodes such that $|\cA_r| \geq P$\;
    $\cA_{r+1} \leftarrow {\rm \texttt{simulate-async}}()$\;
    \For{$i \in \cA_r$}{
        $\wx_i^{(r+1)} \leftarrow \wx_i^{(r)} + \bcC(\bDelta_{\x_i}^{(r)})$\;
        $\wu_i^{(r+1)} \leftarrow \wu_i^{(r)} + \bcC(\bDelta_{\u_i}^{(r)})$\;
        $d_i \leftarrow 0$\;
    }
    \For{$i \in \cV \backslash \cA_r$}{
        \If{$d_i = \tau-1$}{
            $\cA_{r+1} \leftarrow \cA_{r+1} \cup \{i \}$\;
        }
        $\wx_i^{(r+1)} \leftarrow \wx_i^{(r)}$\;
        $\wu_i^{(r+1)} \leftarrow \wu_i^{(r)}$\;
        $d_i \leftarrow d_i + 1$\;
    }
    $ \z^{(r+1)} \leftarrow \arg \min_{\z} ~ h(\z) + \frac{\rho}{2} \sum_{i=1}^N \| \wx_i^{(r+1)} - \z + \wu_i^{(r+1)}\|_2^2$\;
    Broadcast $\bcC(\bDelta_z^{(r)})$ to the nodes\;
    $ r \leftarrow r+1$\;
}
\KwOut{$z^{(r)}, \{\x_{i}^{(r)}\}_{i=1}^N$}
\caption{\texttt{QADMM}}
\end{algorithm}

\end{document}